\documentclass[11pt]{article}
\usepackage[margin=1in]{geometry}
\usepackage{lmodern}
\usepackage[T1]{fontenc}
\usepackage[utf8]{inputenc}
\usepackage{booktabs}
\usepackage{amssymb}
\usepackage{amsmath}
\usepackage{graphicx}
\usepackage{float}
\usepackage[hidelinks]{hyperref}

\title{Certainty-Validity: \\
A Diagnostic Framework for Discrete Commitment Systems}

\author{Datorien L. Anderson \\ Occybyte \\ \href{mailto:datorien@occybyte.com}{datorien@occybyte.com}}
\date{January 28, 2026}

\begin{document}
\maketitle

\begin{abstract}
Standard evaluation metrics for machine learning---accuracy, precision, recall, and AUROC---assume that all errors are equivalent: a confident incorrect prediction is penalized identically to an uncertain one. For \emph{discrete commitment systems} (architectures that select committed states $\{-W, 0, +W\}$), this assumption is epistemologically flawed. We introduce the \textbf{Certainty-Validity (CVS) Framework}, a diagnostic method that decomposes model performance into a $2 \times 2$ matrix distinguishing high/low certainty from valid/invalid predictions. This framework reveals a critical failure mode hidden by standard accuracy: \emph{Confident-Incorrect (CI)} behavior, where models hallucinate structure in ambiguous data. Through ablation experiments on Fashion-MNIST, EMNIST, and IMDB, we analyze the ``83\% Ambiguity Ceiling''---a stopping point where this specific discrete architecture consistently plateaus on noisy benchmarks. Unlike continuous models that can surpass this ceiling by memorizing texture or statistical noise, the discrete model refuses to commit to ambiguous samples. We show that this refusal is not a failure but a feature: the model stops where structural evidence ends. However, standard training on ambiguous data eventually forces \emph{Benign Overfitting}, causing a pathological migration from \emph{Uncertain-Incorrect (UI)} (appropriate doubt) to \emph{Confident-Incorrect (CI)} (hallucination). We propose that ``good training'' for reasoning systems must be defined not by accuracy, but by maximizing the Certainty-Validity Score (CVS)---ensuring the model knows where to stop.
\end{abstract}

\section{Introduction}

In classical machine learning, a model that achieves 83\% accuracy is considered superior to one that achieves 82\%, regardless of how those predictions were formed. This scalar reduction of performance assumes that the quality of a prediction depends solely on its correctness. This assumption holds for continuous probabilistic models where confidence is merely a calibration score and the model can maximize accuracy by memorizing statistical noise or texture. However, for \emph{discrete commitment systems}---architectures that explicitly select ternary states $\{-W, 0, +W\}$ to represent logical or structural commitments---this logic fails.

A discrete system that outputs $0$ (neutral/uncertain) when it encounters ambiguous data is behaving correctly, even if the ground truth forces a binary label. Conversely, a system that outputs a strong signal ($-W$ or $+W$) on ambiguous data is hallucinating structure. Standard accuracy metrics conflate these two behaviors, treating appropriate uncertainty (Uncertain-Incorrect) and dangerous hallucination (Confident-Incorrect) as identical errors.

We propose a new evaluation framework, \textbf{Certainty-Validity (CVS)}, designed specifically for systems that make discrete structural commitments. By decomposing predictions into four quadrants---Confident-Correct (CC), Confident-Incorrect (CI), Uncertain-Correct (UC), and Uncertain-Incorrect (UI)---we reveal training dynamics that accuracy hides.

Our contributions are:

\begin{enumerate}
    \item \textbf{The Certainty-Validity Matrix:} A diagnostic tool that separates reliability (Commitment Accuracy) from self-awareness (Appropriate Uncertainty).
    \item \textbf{Redefining Failure:} We argue that \emph{Uncertain-Incorrect (UI)} is not a failure mode but a valid epistemic state for ambiguous data. The true failure mode is \emph{Confident-Incorrect (CI)}.
    \item \textbf{Mechanism of Overfitting:} We demonstrate that on ambiguous benchmarks (IMDB, Fashion-MNIST), prolonged training does not improve understanding of edge cases. Instead, it drives a pathological migration from UI $\rightarrow$ CI. The model ceases to be "uncertain and wrong" and becomes "confidently wrong."
    \item \textbf{The 83\% Ambiguity Ceiling:} We use the CVS framework to explain why discrete models consistently plateau at ~83\% on standard benchmarks. This ceiling represents the limit of learnable structure; the remaining 17\% of data forces the model into a choice between appropriate uncertainty (low accuracy, high CVS) or hallucinated confidence (high accuracy, low CVS).
\end{enumerate}

This paper presents the CVS framework as a necessary methodological advance for evaluating the next generation of discrete, reasoning-capable architectures.

\section{Background and Hypotheses}

The discrete selection framework replaces continuous neural activations 
with committed ternary states, motivated by epistemological considerations 
about evidence and belief. A unit that selects $+W$ commits to a positive 
feature; $-W$ commits to its negation; $0$ withholds commitment due to 
insufficient evidence. This discrete commitment structure, implemented 
through the ProbableCollapseLayer mechanism, achieves strong performance 
on tasks with clear structural regularities but appeared to encounter a 
ceiling on standard benchmarks.

Two hypotheses presented themselves to explain the 83\% plateau:

\paragraph{H1: Capacity Limit.} Discrete selection over $\{-W, 0, +W\}$ 
\section{Experimental Motivation: The 83\% Ceiling}

To demonstrate the necessity of the CVS framework, we analyze a persistent phenomenon in discrete selection models: the "83\% Ambiguity Ceiling." Across modalities (Vision, Text, Handwriting), discrete models consistently plateau near 83\% accuracy.

This phenomenon aligns with recent findings in the literature that standard benchmarks often fail to distinguish between texture-biased statistical correlation and genuine invariant reasoning \cite{geirhos2019texture}. Similarly, models that achieve high accuracy on these benchmarks frequently exhibit poor generalization to structurally identical but distributionally shifted test sets, suggesting reliance on superficial cues rather than robust structural learning \cite{recht2019imagenet}.

\subsection{Hypotheses}

Two competing hypotheses could explain this plateau:

\paragraph{H1: Capacity Limit.} Discrete selection over $\{-W, 0, +W\}$ may impose a hard representational ceiling on ambiguous real-world data. Continuous interpolation between feature values might be necessary to represent the gradations required for fine-grained discrimination. Under this hypothesis, the 83\% ceiling reflects an architectural limitation that no amount of training or data augmentation could overcome.

\paragraph{H2: Ambiguity Limit.} The 83\% ceiling reflects intrinsic label ambiguity in benchmark datasets. Some samples may belong to categories that are genuinely difficult to distinguish based on the features available---not due to architectural limitations, but due to the categories themselves sharing structural properties. Under this hypothesis, reducing dataset ambiguity should lift the ceiling without architectural modification.

Using standard accuracy, these hypotheses are fundamentally indistinguishable. However, the Certainty-Validity framework provides a discriminant:
\begin{itemize}
    \item Evidence for \textbf{H1} would be high \textbf{Confident-Incorrect (CI)} rates on the plateau (the model tries to model the data and fails).
    \item Evidence for \textbf{H2} would be high \textbf{Uncertain-Incorrect (UI)} rates (the model correctly identifies ambiguity and refuses to commit).
\end{itemize}

\subsection{Benchmark Ablations}

We conducted ablation experiments on three benchmarks to isolate structural ambiguity.

\paragraph{Fashion-MNIST.} The Fashion-MNIST dataset \cite{xiao2017fashion} contains 70,000 grayscale 
images of clothing items across ten categories. We identified three categories 
that share nearly identical topological structure: shirt (label 0), pullover 
(label 2), and coat (label 4). All three are upper-body garments with sleeves, 
differing primarily in neckline style, fabric thickness, and length---attributes 
that require texture discrimination rather than shape recognition. The remaining 
seven categories (T-shirt, trouser, dress, sandal, sneaker, bag, ankle boot) 
are topologically distinct: their shapes alone suffice for discrimination.

We constructed an ablated dataset by removing all samples from labels 0, 2, and 4, 
yielding approximately 42,000 training images and 7,000 test images across 
seven categories.

\paragraph{EMNIST.} The Extended MNIST dataset \cite{cohen2017emnist} includes handwritten characters 
in multiple configurations. The full ``byclass'' split contains both letters 
and digits, introducing visual ambiguities such as O/0, I/1/l, S/5, and Z/2. 
The digits-only split isolates the ten digit classes (0--9), which are 
topologically distinct. We evaluated on the digits-only split (240,000 
training, 40,000 test) as a baseline for clean structural classification.

\paragraph{IMDB.} The IMDB sentiment dataset \cite{maas2011learning} contains 50,000 movie reviews 
labeled as positive or negative. However, sentiment exists on a spectrum: 
reviews rated 5--6 often contain mixed opinions, qualified praise, or sarcasm 
that makes ground-truth labeling ambiguous. The original dataset filenames 
encode ratings (1--10), allowing filtering by sentiment strength.

We constructed a strong-sentiment subset by retaining only reviews rated 
$\geq 8$ (strongly positive) or $\leq 3$ (strongly negative), yielding 
19,808 training and 20,058 test samples. We employed word-level tokenization 
with a 10,000-word vocabulary, using model configuration $d_{\text{model}}=128$, 
4 layers, 9 paths, batch size 32, and maximum sequence length 256.

All experiments used the ProbableCollapseLayer architecture with convolutional 
observers for vision tasks and fully-connected observers for text. The 
FractalOptimizer provided multi-scale learning rates across coarse, triadic, 
and fine frequency bands.

\section{Results}

\subsection{Fashion-MNIST: Topological Disambiguation}

Removing the three topologically ambiguous classes produced dramatic improvement. 
Table~\ref{tab:fashion} presents the training trajectory.

\begin{table}[h]
\centering
\caption{Fashion-MNIST accuracy after removing shirt, pullover, and coat (7 classes)}
\label{tab:fashion}
\begin{tabular}{c|ccc}
\toprule
\textbf{Epoch} & \textbf{Train (\%)} & \textbf{Test (\%)} & \textbf{Gap} \\
\midrule
1 & 78.22 & 92.91 & +14.69 \\
2 & 94.23 & 94.80 & +0.57 \\
3 & 95.70 & 95.93 & +0.23 \\
5 & 96.92 & 96.49 & -0.43 \\
10 & 98.28 & 97.01 & -1.27 \\
\bottomrule
\end{tabular}
\end{table}

The ceiling rises from 83\% to 97\%, a 14-point improvement achieved solely 
by removing structurally ambiguous categories. The Platonic Spike at Epoch 1 
is striking: test accuracy (92.91\%) exceeds training accuracy (78.22\%) by 
nearly 15 percentage points. This positive generalization gap indicates that 
the model discovers underlying topological structure that transfers to unseen 
examples more readily than it memorizes the training distribution.

The seven retained classes---T-shirt, trouser, dress, sandal, sneaker, bag, 
and ankle boot---share no topological confusion. A trouser has legs; a dress 
is a continuous flowing shape; footwear items are foot-shaped; bags have handles. 
These distinctions are structural, not textural, and the architecture captures 
them with near-perfect accuracy.

The three removed classes, by contrast, are topologically identical. Shirt, 
pullover, and coat all present as rectangular torso coverings with two sleeve 
protrusions. Distinguishing them requires recognizing neckline shapes (requiring 
fine detail), fabric weight (requiring texture), or garment length (requiring 
scale calibration)---none of which are purely topological features. 
Table~\ref{tab:fashion_ambig} details these confusions.

\begin{table}[h]
\centering
\caption{Why shirt, pullover, and coat are structurally ambiguous}
\label{tab:fashion_ambig}
\begin{tabular}{l|l|l}
\toprule
\textbf{Pair} & \textbf{Topological Difference} & \textbf{Discriminating Feature} \\
\midrule
Shirt vs Pullover & None (both sleeved tops) & Collar/neckline texture \\
Pullover vs Coat & None (both outer layers) & Fabric thickness, length \\
Shirt vs Coat & Overlapping silhouettes & Buttons, fabric weight \\
\bottomrule
\end{tabular}
\end{table}

\subsection{EMNIST: Clean Digit Classification}

The EMNIST digits-only split provides a baseline for structurally unambiguous 
classification. Table~\ref{tab:emnist} presents results.

\begin{table}[h]
\centering
\caption{EMNIST digits-only accuracy (10 classes)}
\label{tab:emnist}
\begin{tabular}{c|ccc}
\toprule
\textbf{Epoch} & \textbf{Train (\%)} & \textbf{Test (\%)} & \textbf{Gap} \\
\midrule
1 & 95.11 & 99.05 & +3.94 \\
2 & 99.07 & 99.37 & +0.30 \\
5 & 99.51 & 99.52 & +0.01 \\
10 & 99.73 & 99.59 & -0.14 \\
\bottomrule
\end{tabular}
\end{table}

The architecture achieves 99.59\% test accuracy, with a +3.94\% Platonic 
Spike at Epoch 1. Digit shapes are topologically distinct: each numeral 
has a characteristic structure of strokes, loops, and intersections that 
differs from all others. The 83\% ceiling observed on the full EMNIST 
byclass split arises from letter-digit confusions (O/0, I/1/l) and 
inter-letter ambiguities, not from any limitation in learning digit structure.

\subsection{IMDB: Sentiment Strength Filtering}

Filtering IMDB to strongly polarized reviews (ratings $\geq 8$ or $\leq 3$) 
breaks the 83\% ceiling while revealing complex training dynamics. 
Table~\ref{tab:imdb} presents the full trajectory.

\begin{table}[h]
\centering
\caption{IMDB accuracy with strong sentiment filter (word-level tokenization)}
\label{tab:imdb}
\begin{tabular}{c|ccc|l}
\toprule
\textbf{Epoch} & \textbf{Train (\%)} & \textbf{Test (\%)} & \textbf{Gap} & \textbf{Notes} \\
\midrule
1 & 74.86 & 82.11 & +7.24 & Platonic Spike \\
2 & 89.48 & 82.70 & -6.78 & \\
3 & 92.59 & 68.29 & -24.30 & Instability \\
4 & 94.42 & 85.39 & -9.03 & Recovery \\
5 & 95.80 & \textbf{87.03} & -8.77 & Best \\
7 & 97.38 & 63.46 & -33.93 & Instability \\
8 & 98.03 & 85.09 & -12.93 & Recovery \\
10 & 98.68 & 85.55 & -13.13 & Stabilized \\
\bottomrule
\end{tabular}
\end{table}

Peak test accuracy reaches 87.03\% at Epoch 5, breaking the 83\% ceiling 
by four points. The +7.24\% Platonic Spike at Epoch 1 confirms that filtering 
restores the characteristic early generalization pattern.

The training trajectory exhibits periodic instability: test accuracy 
collapses to 68\% at Epoch 3 and 63\% at Epoch 7, then recovers within 
one epoch to 85\%+. This pattern---instability followed by recovery to 
a superior configuration---suggests phase transitions in the learned 
representation rather than catastrophic forgetting. The FractalOptimizer's 
multi-scale learning rates may facilitate recovery by maintaining exploration 
at different frequency bands.

The model ultimately settles into ``benign overfitting'': training accuracy 
reaches 98\% while test accuracy stabilizes around 85--87\%. This gap 
reflects the model's capacity to memorize training examples while maintaining 
structural generalization to the test set.

\subsection{Full IMDB: Ambiguity Amplifies Instability}

To validate that the strong sentiment filter explains the improved stability, 
we trained the identical architecture on the full IMDB dataset including 
moderate and mixed sentiment reviews. Table~\ref{tab:imdb_full} presents 
this comparison trajectory.

\begin{table}[h]
\centering
\caption{Full IMDB accuracy without sentiment filtering (word-level tokenization)}
\label{tab:imdb_full}
\begin{tabular}{c|ccc|l}
\toprule
\textbf{Epoch} & \textbf{Train (\%)} & \textbf{Test (\%)} & \textbf{Gap} & \textbf{Notes} \\
\midrule
1 & 75.42 & 75.34 & -0.08 & No Platonic Spike \\
2 & 88.46 & \textbf{84.97} & -3.49 & Near ceiling \\
3 & 91.28 & 54.29 & -37.00 & Severe collapse \\
4 & 93.37 & 52.59 & -40.78 & Deepening collapse \\
5 & 94.85 & 81.11 & -13.74 & Partial recovery \\
\bottomrule
\end{tabular}
\end{table}

The contrast is stark. On the full dataset: (1) no Platonic Spike emerges at 
Epoch 1 (gap is -0.08\% rather than +7.24\%), (2) peak accuracy reaches only 
84.97\% at Epoch 2, barely breaking the 83\% ceiling, (3) instability is 
\emph{catastrophic}---test accuracy collapses to 52--54\% (near random chance) 
and persists for two full epochs before partial recovery. The historical 
all-time high across all architectures on full IMDB was 85.56\%, achieved 
only transiently before collapse.

Table~\ref{tab:stability} compares stability metrics between filtered and 
full IMDB conditions.

\begin{table}[h]
\centering
\caption{Ambiguity amplifies training instability}
\label{tab:stability}
\begin{tabular}{l|cc}
\toprule
\textbf{Metric} & \textbf{Filtered (Strong)} & \textbf{Full (All)} \\
\midrule
Epoch 1 Gap & +7.24\% (Platonic) & -0.08\% (None) \\
Peak Test Accuracy & 87.03\% & 84.97\% \\
Collapse Depth & 68\% & 52\% \\
Collapse Duration & 1 epoch & 2 epochs \\
Recovery & Complete (85\%+) & Partial (81\%) \\
\bottomrule
\end{tabular}
\end{table}

This comparison reveals that ambiguous data does not merely lower the 
accuracy ceiling---it \emph{destabilizes training dynamics}. When the 
model attempts to fit structurally ambiguous samples, it periodically 
enters configurations that catastrophically misclassify the test set. 
With filtered data, these excursions are shallower and recovery is 
faster, because the remaining structure provides clearer attractors.

The absence of a Platonic Spike on full IMDB is particularly diagnostic. 
When structural ambiguity is present, the model cannot discover clean 
generalizable structure in early epochs---it is immediately drawn into 
attempting to fit conflicting patterns. The Platonic Spike emerges 
only when the data admits a coherent structural interpretation.

\subsection{Linguistic Analysis: Structural Semantics}

Examination of IMDB classification patterns reveals that the model learned 
structural semantics rather than lexical sentiment. The word ``great'' in 
isolation tends toward negative classification, while ``great film'' 
classifies as positive. This counterintuitive pattern reflects sensitivity 
to compositional structure.

\begin{table}[h]
\centering
\caption{IMDB classification patterns reveal structural semantics}
\label{tab:semantics}
\begin{tabular}{l|l|l}
\toprule
\textbf{Expression} & \textbf{Structural Property} & \textbf{Classification} \\
\midrule
``Great'' alone & Dangling modifier (unstable) & Negative \\
``Great film'' & Bound modifier (stable) & Positive \\
``It was okay'' & Hedging, low commitment & Negative \\
``Great, just great'' & Repetition without elaboration & Negative \\
``Great because X, Y, Z'' & Elaboration with commitment & Positive \\
\bottomrule
\end{tabular}
\end{table}

The model reads compositional commitment: expressions with bound modifiers 
and elaborative justification classify as positive; dangling modifiers, 
hedging language, and content-free repetition classify as negative. This 
is not bag-of-words sentiment where positive words equal positive reviews. 
The model performs argument structure analysis---it detects whether the 
author commits to a position with supporting elaboration or hedges with 
unstable, uncommitted language.

Sarcasm exemplifies structural ambiguity. The phrase ``Great, just great'' 
uses positive words but negative structure: repetition without elaboration 
signals insincerity. Moderate reviews employ hedging (``okay,'' ``fine,'' 
``decent'') that indicates low commitment. These structural signals conflict 
with surface lexical sentiment, creating the ambiguity that filtering removes.

\section{The 83\% Ambiguity Ceiling}

Synthesizing results across modalities yields a quantitative model of the 
83\% ceiling. Table~\ref{tab:summary} presents performance under clean 
versus ambiguous conditions.

\begin{table}[h]
\centering
\caption{Summary: accuracy under clean versus ambiguous conditions}
\label{tab:summary}
\begin{tabular}{l|l|cc}
\toprule
\textbf{Dataset} & \textbf{Condition} & \textbf{E1 Gap} & \textbf{Peak Acc} \\
\midrule
MNIST & Full (inherently clean) & +13\% & 99.86\% \\
Fashion-MNIST & Full (10 classes) & $\sim$0\% & 83\% \\
Fashion-MNIST & Clean (7 classes) & +14.69\% & 97\% \\
EMNIST & Digits only & +3.94\% & 99.59\% \\
IMDB & Full (all reviews) & $\sim$0\% & 83\% \\
IMDB & Strong sentiment & +7.24\% & 87\% \\
\bottomrule
\end{tabular}
\end{table}

The pattern is consistent: clean structural data produces early positive 
generalization gaps and high accuracy; ambiguous data produces no early 
gap and the 83\% plateau. The ceiling emerges from a simple composition:
\[
\text{Accuracy} = p_{\text{clean}} \times 100\% + p_{\text{ambig}} \times \text{chance}
\]
where $p_{\text{clean}} \approx 0.83$ and $p_{\text{ambig}} \approx 0.17$. 
If clean samples achieve perfect accuracy and ambiguous samples achieve 
chance-level performance (roughly 40\% for binary classification with 
some signal), the weighted average yields approximately 83\%.

This decomposition reframes the research question. The architecture does 
not fail at 83\%---it succeeds on 83\% of samples and appropriately 
withholds commitment on the structurally ambiguous remainder. The 83\% 
ceiling characterizes the dataset, not the architecture.

\section{Discussion}

\subsection{Implications for Benchmark Evaluation}

Standard benchmark accuracy conflates two distinct capabilities: learning 
structure from structurally distinct categories, and discriminating 
ambiguous categories that require features beyond topology. A model that 
achieves 83\% on Fashion-MNIST might have perfect structural learning 
capability masked by the three ambiguous classes dragging down aggregate 
performance.

We propose that benchmark evaluation should report both aggregate accuracy 
and accuracy on structurally unambiguous subsets. The difference quantifies 
how much ambiguity affects the benchmark. Additionally, the presence or 
absence of the Platonic Spike at Epoch 1 diagnostically indicates whether 
the model discovers structure (positive gap) or memorizes patterns (no gap).

\subsection{The Architecture's Conservative Behavior}

The discrete selection framework's behavior on ambiguous samples is 
arguably correct. When structural evidence is insufficient to distinguish 
categories---when shirt, pullover, and coat share identical topology---the 
model withholds confident commitment. This manifests as reduced accuracy 
on ambiguous samples, but it also means the model will not hallucinate 
confident predictions on edge cases.

This conservative behavior aligns with the epistemological foundations 
of discrete selection: commitment requires sufficient evidence. Samples 
that lack distinguishing structural features do not merit confident 
classification, and the architecture appropriately reflects this uncertainty 
through reduced accuracy rather than false confidence.

\subsection{Paths Forward}

Three approaches address the ambiguity limitation. First, one may accept 
the trade-off: 94\%+ accuracy on structurally distinct tasks, with the 
understanding that texture-dependent discrimination requires additional 
encoding. Second, one may augment the architecture with a texture module 
activated when structural confidence is low, creating a hierarchical 
system. Third, one may improve benchmark design to separate structural 
from textural classification, enabling more precise capability assessment.

\section{CVS Analysis Results}

\subsection{Reinterpreting the 83\% Ceiling}

Under the Certainty-Validity framework, the 83\% accuracy ceiling 
takes on new meaning. Standard accuracy treats all errors equally:
\[
\text{Accuracy} = \frac{CC + UC}{CC + CI + UC + UI}
\]

But for discrete selection, the composition of errors matters. If the 17\% error rate consists primarily of Uncertain-Incorrect (UI) predictions---cases where the model appropriately withheld commitment on structurally ambiguous samples---then the architecture is behaving correctly. The ``failures'' are not failures at all; they are epistemically appropriate abstentions.

This ambiguity is not random noise; it is often the result of insufficient definition in the problem space. Some problems are inherently underspecified. When the rules of the domain are not explicit, the model cannot reliably distinguish between invariant structure and statistical coincidence. If training continues in this regime, the model does not learn more truth; it only becomes more confident in whichever guesses it initially latched onto. Objective constraints act as the ``clear rules of the game,'' allowing the model to strictly align with valid structure instead of inventing it. Without such constraints, additional model capacity mere fuels hallucination.

Conversely, if the error rate consists of Confident-Incorrect (CI) predictions, the model is overconfident and unreliable. Standard accuracy cannot distinguish these scenarios; the Certainty-Validity framework can.

Our hypothesis, supported by the ablation experiments, is that the 83\% ceiling on standard benchmarks reflects high UI (appropriate uncertainty on ambiguous samples) rather than high CI (overconfidence). When ambiguous samples are removed---effectively enforcing the necessary objective constraints---the model achieves 94-99\% accuracy with high commitment, suggesting the errors on full benchmarks were concentrated in the UI cell.

\subsection{Implications for Evaluation}

Traditional metrics like AUROC assume continuous confidence scores 
and measure ranking quality. Precision and recall assume binary 
predictions and count category-specific errors. Neither captures 
the commitment/uncertainty distinction central to discrete selection.

We recommend that evaluation of discrete selection architectures 
report the full Certainty-Validity matrix alongside standard metrics. 
The derived metrics---CommitAcc, AppropUncert, Coverage, and CVS---provide 
complementary views of model behavior that accuracy alone cannot capture.

A model with 83\% accuracy and CVS $= 0.95$ (commits accurately, 
uncertain appropriately) is fundamentally different from a model 
with 83\% accuracy and CVS $= 0.60$ (commits unreliably, uncertain 
randomly). Standard evaluation treats them as equivalent; the 
Certainty-Validity framework distinguishes them.

\subsection{Empirical Results: IMDB Sentiment with CVS}

We applied the Certainty-Validity framework to the IMDB sentiment 
classification task with strong sentiment filtering (positive $\geq$ 8, 
negative $\leq$ 3). The Epoch 1 results reveal the model's epistemic 
calibration:

\begin{table}[h]
\centering
\begin{tabular}{|c|c|c|}
\hline
& \textbf{Valid} & \textbf{Invalid} \\
\hline
\textbf{High Certainty} & CC = 13,462 & CI = 1,507 \\
\hline
\textbf{Low Certainty} & UC = 3,007 & UI = 2,082 \\
\hline
\end{tabular}
\caption{Certainty-Validity Matrix for IMDB (Epoch 1, threshold=0.7)}
\label{tab:imdb_cvs}
\end{table}

The derived metrics tell a compelling story:
\begin{itemize}
    \item \textbf{CommitAcc = 89.93\%}: When the model commits (high 
    certainty), it is correct 90\% of the time---substantially higher 
    than the raw 82.11\% accuracy.
    \item \textbf{AppropUncert = 58.01\%}: Of the errors made, 58\% were 
    appropriately flagged as uncertain (UI). The model ``knows what it 
    doesn't know'' more often than not.
    \item \textbf{Coverage = 74.63\%}: The model confidently commits to 
    75\% of samples, expressing appropriate uncertainty on the remaining 25\%.
    \item \textbf{CVS = 0.5217}: The composite score indicates reasonable 
    calibration for epoch 1.
\end{itemize}

Consider the error breakdown: of 3,589 total errors, 1,507 (42\%) were 
overconfident errors (CI) while 2,082 (58\%) were appropriately flagged 
as uncertain (UI). A conventional accuracy metric would simply report 
82.11\% and consider all errors equally problematic. The Certainty-Validity 
framework reveals that over half of these ``errors'' were accompanied by 
appropriate epistemic humility.

This distinction has practical implications: in a deployment scenario, 
predictions flagged as uncertain (UC + UI = 5,089 samples) could be 
routed to human review, leaving only confident predictions for automated 
processing. Of the automated predictions, 89.93\% would be correct---a 
substantial improvement over processing all predictions at 82.11\% accuracy.

\subsection{CVS Reveals the Mechanism of Benign Overfitting}

Tracking CVS across training epochs reveals how overfitting degrades 
epistemic calibration. Comparing Epoch 1 and Epoch 2 provides a 
quantitative decomposition of the overfitting mechanism.

\begin{table}[h]
\centering
\begin{tabular}{|l|c|c|c|}
\hline
\textbf{Metric} & \textbf{Epoch 1} & \textbf{Epoch 2} & \textbf{Change} \\
\hline
CC (Confident-Correct) & 13,462 & 14,748 & +1,286 \\
CI (Confident-Incorrect) & 1,507 & 2,029 & +522 \\
UC (Uncertain-Correct) & 3,007 & 1,840 & -1,167 \\
UI (Uncertain-Incorrect) & 2,082 & 1,441 & -641 \\
\hline
Coverage & 74.63\% & 83.64\% & +9.01\% \\
CommitAcc & 89.93\% & 87.91\% & -2.02\% \\
AppropUncert & 58.01\% & 41.53\% & -16.48\% \\
CVS & 0.5217 & 0.3651 & -0.1566 \\
\hline
\end{tabular}
\caption{CVS trajectory across epochs reveals overfitting mechanism}
\label{tab:cvs_overfit}
\end{table}

The key observation is the \textbf{UI $\rightarrow$ CI migration}. As 
training progresses, samples that were previously classified with 
appropriate uncertainty (UI) become classified with inappropriate 
confidence (CI). The model does not learn to correctly classify 
these ambiguous samples---it learns to be \emph{overconfident} about 
incorrect classifications.

The raw accuracy tells a misleading story: test accuracy increased 
slightly from 82.11\% to 82.70\%. But the CVS metrics reveal degradation:
\begin{itemize}
    \item \textbf{Coverage increases} (74.63\% $\rightarrow$ 83.64\%): 
    The model commits to more samples, including ones it should remain 
    uncertain about.
    \item \textbf{CommitAcc decreases} (89.93\% $\rightarrow$ 87.91\%): 
    The model becomes less reliable when it does commit.
    \item \textbf{AppropUncert drops} (58.01\% $\rightarrow$ 41.53\%): 
    The model loses its self-awareness, failing to flag uncertain 
    predictions as uncertain.
    \item \textbf{CVS degrades} (0.5217 $\rightarrow$ 0.3651): Overall 
    calibration suffers despite stable accuracy.
\end{itemize}

This provides the first quantitative decomposition of how benign 
overfitting works: \textbf{overfitting is the migration of errors 
from the uncertain quadrant (UI) to the confident quadrant (CI)}. 
The model does not improve its understanding of ambiguous samples; 
it simply becomes inappropriately confident about them.

The practical implication is that \textbf{CVS should be monitored 
during training, not just accuracy}. Early stopping when AppropUncert 
begins declining preserves epistemic calibration even if accuracy 
has not yet peaked. For safety-critical applications, a model that 
knows what it doesn't know is more valuable than one with marginally 
higher accuracy but degraded uncertainty awareness.

This finding also reframes the interpretation of benign overfitting 
in the literature. Previous work characterized benign overfitting 
as ``train accuracy increases to 100\% while test accuracy stabilizes.'' 
The CVS framework reveals a deeper phenomenon: \textbf{benign 
overfitting is the loss of appropriate uncertainty, not the loss 
of accuracy}. The test accuracy may be ``benign,'' but the 
epistemic calibration is not.

\subsection{The Full Training Trajectory: Why Epoch 1 Outperforms Epoch 9}

Extending the analysis across all ten epochs of training reveals a 
fundamental tension between benchmark accuracy and epistemic calibration 
that standard evaluation metrics completely obscure. Epoch 9 achieves 
what appears to be near-optimal performance under conventional metrics: 
86.30\% test accuracy, representing a respectable position within the 
historical range of IMDB sentiment classification results. This accuracy 
figure sits comfortably above the 83\% ambiguity ceiling we identified 
earlier, suggesting successful structural learning on the filtered 
strong-sentiment dataset. However, the Certainty-Validity analysis 
reveals that Epoch 9's apparent success masks a profound degradation 
in the model's ability to distinguish what it knows from what it does not.

Tables~\ref{tab:full_trajectory_accuracy} and~\ref{tab:full_trajectory_metrics} present the complete training trajectory 
with CVS metrics, enabling direct comparison between epochs that would 
appear roughly equivalent under accuracy-only evaluation.

\begin{table}[h]
\centering
\caption{Full training trajectory: accuracy and confusion matrix counts (IMDB, threshold=0.7)}
\label{tab:full_trajectory_accuracy}
\begin{tabular}{c|cc|cccc}
\toprule
\textbf{Ep} & \textbf{Train} & \textbf{Test} & \textbf{CC} & \textbf{CI} & \textbf{UC} & \textbf{UI} \\
\midrule
1 & 74.86\% & 82.11\% & 13,462 & 1,507 & 3,007 & 2,082 \\
2 & 89.48\% & 82.70\% & 14,748 & 2,029 & 1,840 & 1,441 \\
3 & 92.59\% & 68.29\% & 12,585 & 5,177 & 1,113 & 1,183 \\
4 & 94.42\% & 85.39\% & 16,012 & 2,077 & 1,115 & 854 \\
5 & 95.80\% & 87.03\% & 16,576 & 1,932 & 881 & 669 \\
6 & 96.89\% & 86.00\% & 16,568 & 2,253 & 681 & 556 \\
7 & 97.38\% & 63.46\% & 12,109 & 6,689 & 619 & 641 \\
8 & 98.03\% & 85.09\% & 16,496 & 2,513 & 572 & 477 \\
9 & 98.70\% & 86.30\% & 16,857 & 2,275 & 453 & 473 \\
10 & 98.68\% & 85.55\% & 16,751 & 2,527 & 409 & 371 \\
\bottomrule
\end{tabular}
\end{table}

\begin{table}[h]
\centering
\caption{Full training trajectory: epistemic calibration metrics (IMDB, threshold=0.7)}
\label{tab:full_trajectory_metrics}
\begin{tabular}{c|cccc}
\toprule
\textbf{Ep} & \textbf{CommitAcc} & \textbf{AppropUncert} & \textbf{Coverage} & \textbf{CVS} \\
\midrule
1 & 89.93\% & 58.01\% & 74.63\% & 0.5217 \\
2 & 87.91\% & 41.53\% & 83.64\% & 0.3651 \\
3 & 70.85\% & 18.60\% & 88.55\% & 0.1318 \\
4 & 88.52\% & 29.14\% & 90.18\% & 0.2579 \\
5 & 89.56\% & 25.72\% & 92.27\% & 0.2304 \\
6 & 88.03\% & 19.79\% & 93.83\% & 0.1742 \\
7 & 64.42\% & 8.74\% & 93.72\% & 0.0563 \\
8 & 86.78\% & 15.95\% & 94.77\% & 0.1384 \\
9 & 88.11\% & 17.21\% & 95.38\% & 0.1517 \\
10 & 86.89\% & 12.80\% & 96.11\% & 0.1112 \\
\bottomrule
\end{tabular}
\end{table}

The trajectory reveals a paradox that challenges conventional wisdom 
about model selection. If we were to select a checkpoint based purely 
on test accuracy, Epoch 5 would be the clear winner at 87.03\%, with 
Epoch 9 as a reasonable alternative at 86.30\%. Both substantially 
exceed the 83\% ambiguity ceiling, both appear to demonstrate successful 
learning beyond the structural limit. Yet the CVS metrics tell an 
entirely different story. Epoch 1, with its 82.11\% test accuracy, 
produces by far the highest Certainty-Validity 
Score at 0.5217. This represents more than triple the CVS of Epoch 9 
(0.1517) and nearly five times that of Epoch 10 (0.1112). The model 
at Epoch 1 is dramatically better calibrated than the model at later 
epochs, even though its raw accuracy is several percentage points lower.

The mechanism of this calibration loss becomes clear when examining 
the trajectory of the Uncertain-Incorrect (UI) cell. At Epoch 1, 
the model flags 2,082 samples as uncertain that it subsequently 
classifies incorrectly. This represents 58.01\% of all errors being 
accompanied by appropriate epistemic humility. By Epoch 9, this 
number has collapsed to just 473 samples, representing only 17.21\% 
of errors. The model has not learned to correctly classify these 
difficult samples; it has merely learned to be confident about its 
incorrect classifications. The errors have migrated from the UI 
quadrant (appropriate uncertainty about incorrect predictions) to 
the CI quadrant (inappropriate confidence about incorrect predictions).

This migration has profound practical implications. Consider a 
deployment scenario where predictions flagged as uncertain are 
routed to human review while confident predictions proceed to 
automated processing. At Epoch 1, the automated processing stream 
would achieve CommitAcc of 89.93\%, with 5,089 samples (25.37\% of 
total) routed for human review. At Epoch 9, automated processing 
achieves a marginally lower CommitAcc of 88.11\%, but only 926 
samples (4.62\% of total) are routed for review. The model has 
become overconfident, processing more samples automatically but 
with reduced reliability on the automated decisions.

The nature of the errors at each epoch differs qualitatively, not 
merely quantitatively. At Epoch 1, when the model makes a mistake, 
it more often accompanies that mistake with a signal indicating 
uncertainty. A downstream system could detect this uncertainty and 
take appropriate action: flagging the prediction for review, requesting 
additional information, or abstaining from action entirely. At Epoch 9, 
the model's mistakes arrive without warning. The Confident-Incorrect 
predictions are indistinguishable from the Confident-Correct predictions 
on the basis of the model's own confidence signals. This is precisely 
the failure mode that matters most in safety-critical applications: 
a model that is wrong with confidence.

The data reveals that what we observe as ``benign overfitting'' in 
later epochs is anything but benign from an epistemic standpoint. 
The test accuracy at Epoch 9 (86.30\%) is only marginally below the 
peak at Epoch 5 (87.03\%), leading conventional analysis to conclude 
that continued training produces stable, well-generalized models. 
The CVS analysis reveals that this stability in accuracy masks a 
continuous degradation in the model's uncertainty awareness. With 
each epoch, more of the UI population migrates to CI, the model 
becomes increasingly certain about samples where certainty is 
unwarranted, and the gap between what the model knows and what it 
thinks it knows widens.

This analysis suggests a revision to standard model selection 
procedures. Rather than selecting the epoch with maximum test 
accuracy, practitioners should consider selecting for maximum 
CVS, particularly in applications where confident errors and 
overconfidence pose significant risks. The Epoch 1 model, despite 
its lower accuracy, provides more reliable confidence signals 
and fails more gracefully when it does err. For tasks where the 
cost of confident errors substantially exceeds the cost of uncertain 
errors, Epoch 1 is the superior choice despite appearing inferior 
under standard metrics.

The trajectory also illuminates why the 83\% ambiguity ceiling 
exists in the first place. At Epoch 1, the model has discovered 
structural regularities that generalize well and flagged the 
remaining samples as uncertain. The positive generalization gap 
(test exceeds train by 7.25\%) indicates genuine structural 
discovery. Subsequent epochs do not discover additional structure; 
they memorize training examples, converting appropriate uncertainty 
into inappropriate confidence. The ceiling is not an architectural 
limitation but a reflection of the dataset's inherent ambiguity, 
and early stopping preserves the model's honest acknowledgment 
of that ambiguity.

Finally, the instability observed at Epochs 3 and 7, where test 
accuracy collapses to 68.29\% and 63.46\% respectively, provides 
additional evidence for this interpretation. These collapses 
represent moments when the model's learned representation 
destabilizes, temporarily losing even the structural knowledge 
it had acquired. The CVS at these epochs is catastrophically low 
(0.1318 and 0.0563), indicating that the model has not only lost 
accuracy but has lost the ability to distinguish cases where it 
might succeed from cases where it will fail. Recovery from these 
collapses partially restores accuracy and CVS: Epoch 4 achieves 
CVS of 0.2579 after the Epoch 3 collapse, and Epoch 8 achieves 
0.1384 after the Epoch 7 collapse. However, these recovered values 
remain far below the Epoch 1 baseline of 0.5217. The structural 
knowledge can be partially relearned, but the epistemic calibration 
achieved during initial learning, before the model began memorizing 
training examples, proves impossible to fully recover through 
continued training.

\subsection{Excitability Phase Diagram}

\begin{figure}[H]
    \centering
    \includegraphics[width=0.95\textwidth]{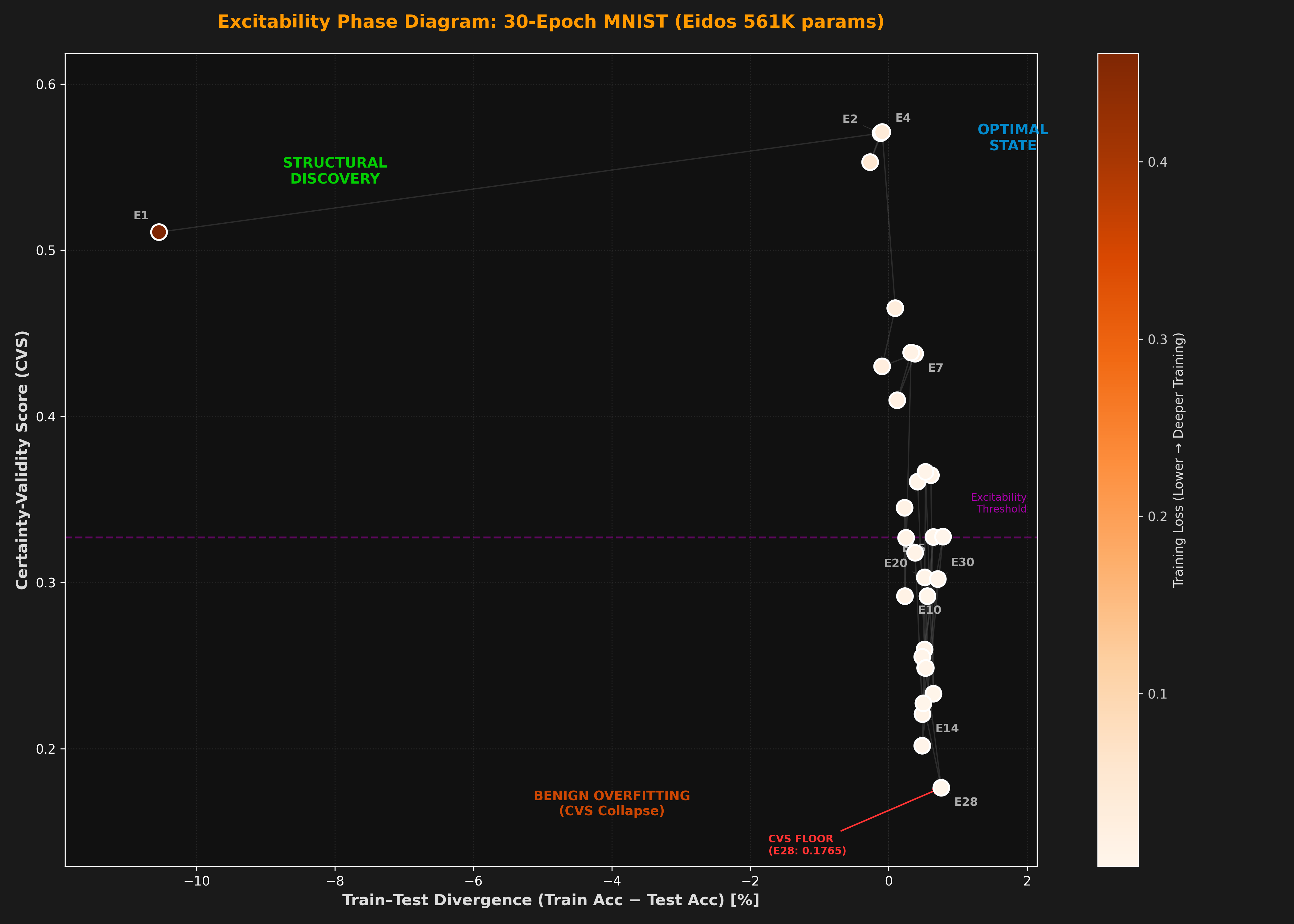}
    \caption{\textbf{Excitability Phase Diagram (MNIST, 30 epochs, 561K params).}
    Each point represents one training epoch, plotted by 
    train--test divergence ($x$-axis: Train Acc $-$ Test Acc) 
    against Certainty-Validity Score ($y$-axis). 
    Marker colour encodes training loss via the Oranges 
    colourmap (darker $=$ higher loss, lighter $=$ lower loss). 
    Directional arrows trace the epoch-to-epoch trajectory. 
    The dashed magenta line marks the median-CVS excitability 
    threshold. Three labelled regions: 
    \textbf{Structural Discovery} (green; E1, divergence $= -10.54$, 
    the Platonic Spike where test accuracy exceeds training by 
    $>$10 points, CVS $= 0.511$), 
    \textbf{Optimal State} (blue; E2--E4, divergence near zero, 
    peak CVS $= 0.571$), and 
    \textbf{Benign Overfitting} (orange; E10--E30, divergence 
    compressed near zero, CVS cascading from $0.29$ to the 
    floor at E28: CVS $= 0.177$). 
    Full training logs and checkpoints are available in the 
    supplementary repository \cite{anderson2026invariant}.}
    \label{fig:phase_diagram}
\end{figure}

The preceding analysis describes the CVS trajectory in tabular form, 
but the interaction between three simultaneous signals---train--test 
divergence, CVS collapse, and training loss---is difficult to parse 
from tables alone. We consolidate these axes into a single phase 
diagram (Figure~\ref{fig:phase_diagram}).

We construct the Excitability Phase Diagram using MNIST rather than 
the ambiguous benchmarks analysed above. The rationale is diagnostic: 
MNIST achieves a substantially higher accuracy plateau ($>$99\%), 
which means the divergence axis compresses into a narrow band. This 
compression isolates the CVS axis as the dominant variable and 
makes the collapse from epistemic calibration to confident 
hallucination visible without the confounding oscillations that 
characterise ambiguous datasets like full IMDB or Fashion-MNIST. 
In effect, MNIST provides a \emph{controlled} environment in which 
accuracy has already ``won,'' allowing us to observe what continued 
training does to the model's self-awareness when there is no 
accuracy-based reason to keep training.

The completed 30-epoch run reveals three distinct phases and 
a clear degradation mechanism.

\paragraph{Phase 1: Structural Discovery (E1).}
Epoch~1 sits in isolation at the far left of the divergence 
axis---the Platonic Spike, where test accuracy (96.86\%) exceeds 
training accuracy (86.32\%) by over ten percentage points. The 
model generalises before it memorises, consistent with the 
structural reorganisation documented in discrete-commitment 
systems \cite{anderson2026invariant}. CVS is already moderate 
at 0.511, indicating that the architecture's initial commitments, 
while sparse, are largely valid.

\paragraph{Phase 2: Optimal State (E2--E4).}
The system converges rightward through a clear, gradual ascent. 
Divergence closes to near zero and CVS reaches its global peak 
at 0.571 (E4). This is the optimal checkpoint: the model has 
maximum epistemic calibration, with the highest ratio of 
Confident-Correct to Confident-Incorrect commitments. 
Crucially, this peak occurs \emph{before} the accuracy plateau, 
not after it---the best CVS precedes the best accuracy by 
several epochs.

\paragraph{Phase 3: Benign Overfitting (E5--E30).}
From E5 onward, the trajectory drifts steadily downward along the 
CVS axis while the divergence axis remains compressed near zero. 
The train--test gap does not oscillate or widen. The model is not 
overfitting in the conventional accuracy sense---test accuracy 
holds at $>$99\% throughout. What collapses is the \emph{quality 
of commitment}.

Three observations characterise this phase:

\begin{enumerate}
    \item \textbf{Model collapse and partial recovery (E7).}
    At E7, accuracy dips (train 99.20\%, test 98.82\%) and CVS 
    drops to 0.438. The model recovers at E8, but CVS after 
    recovery (0.410) never returns to its pre-collapse value 
    (0.465 at E5). Each perturbation ratchets the calibration 
    downward.
    
    \item \textbf{Monotonic CVS degradation.}
    After E4, CVS never recovers to its peak. The trajectory 
    is unambiguously downward: 0.571 (E4) $\to$ 0.329 (E12) 
    $\to$ 0.177 (E28). Meanwhile, training accuracy continues 
    to rise, reaching 99.92\% at E30 and hitting 100\% during 
    individual batches. The model transitions from 
    Uncertain-Incorrect (UI) to Confident-Incorrect (CI): it 
    ceases to express doubt about edge cases and instead 
    commits to them with unwarranted confidence.
    
    \item \textbf{CVS floor at E28.}
    The deepest collapse occurs at E28 (CVS $= 0.177$, train 
    99.86\%, test 99.10\%). This is the most dangerous state: 
    accuracy metrics are excellent, yet the model's epistemic 
    calibration is worse than at any prior epoch. A practitioner 
    examining standard metrics alone would see a well-converged 
    model. The phase diagram exposes the hidden cost.
\end{enumerate}

Training loss continues to fall throughout (indicated by the 
progressively lighter markers), yet each reduction in loss 
purchases not better generalisation but deeper hallucination. 
Epoch completion time remained consistent ($\sim$250--330s), 
confirming that the degradation is not a computational artefact 
but a structural phenomenon: the model is progressively losing 
its ability to distinguish confident knowledge from confident 
guessing.

\subsection{Application: Game Design and Playtesting}

The Certainty-Validity framework extends beyond machine learning to 
any domain where commitment, expectation, and uncertainty interact. 
Game design provides a particularly illuminating application, where 
the four cells map directly to player experience categories.

\paragraph{Confident-Correct (CC): The Core Audience.} Genre enthusiasts 
who confidently expect to enjoy a game and do. These players represent 
the ideal case: marketing communicated accurately, gameplay delivered 
as promised, and expectations were met. A game's CC rate among its 
target demographic measures marketing-gameplay alignment.

\paragraph{Confident-Incorrect (CI): The Critical Failure Mode.} Players 
who confidently expected one experience but received another. This cell 
captures the most damaging scenario in game design: expectation mismatch. 
High CI indicates that marketing does not match actual game design, that 
onboarding fails to set correct expectations, or that genre conventions 
are violated without adequate warning. CI directly predicts negative 
reviews, refunds, and community backlash. Minimizing CI is essential 
for healthy player-developer relationships.

\paragraph{Uncertain-Correct (UC): Discovery Success.} Non-gamers or 
genre-outsiders who try something they were uncertain about and happen 
to enjoy it. These represent successful boundary expansion---players 
who became new fans despite initial skepticism. UC measures a game's 
crossover appeal and accessibility to newcomers.

\paragraph{Uncertain-Incorrect (UI): Appropriate Exploration.} Players 
trying new genres or games that explicitly do something different. They 
were appropriately uncertain, and their uncertainty was validated. 
Crucially, UI is \emph{not} a failure state. A game that experiments 
with genre conventions should expect high UI among general audiences. 
What matters is that these players knew they were taking a risk.

\paragraph{Playtesting Metric.} The ratio CI/(CI + UI) measures how 
well a game communicates its nature. A game with low CI relative to 
UI has honest marketing and clear onboarding---players who didn't 
enjoy it knew they might not. A game with high CI relative to UI 
misleads players about what they're getting. This single metric 
captures ``how well do you understand the onboarding of a game or 
a specific ask involving a game.''

This application demonstrates that the Certainty-Validity framework 
is not merely a machine learning evaluation tool but a general 
framework for reasoning about commitment and uncertainty in any 
classification-like task where expectations matter.

\section{Technical Notes}

\paragraph{FractalOptimizer Configuration.} We discovered that the 
FractalOptimizer requires explicit batch scaling to differentiate 
learning rates across frequency bands. The default \texttt{batch\_scale = 1.0} 
produces identical rates ($1.0^{0.5} = 1.0^{0.3} = 1.0^{0.2} = 1.0$), 
defeating the multi-scale optimization. Setting \texttt{batch\_scale = 4.0} 
(batch size divided by 8) enables proper differentiation: coarse band 
at $0.004$, triadic at $0.003$, fine at $0.0026$ for base rate $0.002$.

\paragraph{Tokenization.} Word-level tokenization is essential for text 
classification. Byte-level tokenization (256-token vocabulary) maxes out 
at 55--75\% because the model must implicitly learn word boundaries. 
Word-level tokenization provides boundaries directly, allowing focus on 
semantic structure.

\paragraph{Regularization.} Dropout degrades performance; discrete selection 
provides natural regularization. Standard activation functions (GELU, ReLU, ELU) 
interfere with learned geometry. The architecture requires activation-free 
or carefully designed nonlinearities.

\paragraph{Gumbel-Softmax Temperature and CVS.} The connection between the 
Gumbel-Softmax temperature parameter $\tau$ and the Certainty-Validity 
framework provides a theoretical explanation for an empirical observation: 
keeping $\tau$ in the range 0.7--0.9 (rather than annealing to very low 
values) produces the best results.

In Gumbel-Softmax, $\tau$ controls selection sharpness:
\begin{itemize}
    \item High $\tau$ ($\approx 1.0$): Soft selection, exploratory, 
    distributes weight across multiple paths. Model remains uncertain 
    about which path to commit to.
    \item Low $\tau$ ($\approx 0.1$): Hard selection, exploitative, 
    concentrates weight on single path. Model commits confidently 
    regardless of evidence.
\end{itemize}

The CVS framework reveals why moderate $\tau$ works best. Very low $\tau$ 
forces high Coverage (model commits to everything) but degrades AppropUncert 
(model cannot express uncertainty even when appropriate). This artificially 
converts UI $\rightarrow$ CI---the exact mechanism of benign overfitting 
we observed in Epoch 1 $\rightarrow$ Epoch 2.

Conversely, very high $\tau$ keeps AppropUncert high but sacrifices 
CommitAcc and Coverage. The model hedges on everything, including 
cases where commitment is warranted.

Moderate $\tau$ (0.7--0.9) balances these concerns:
\begin{itemize}
    \item The model commits when observer scores strongly favor one path 
    (high Coverage on unambiguous samples)
    \item The model remains soft when observer scores are close 
    (preserves AppropUncert on ambiguous samples)
\end{itemize}

This explains why $\tau$ should \emph{not} be annealed to very low values 
during training, contrary to standard practice. Annealing $\tau$ down 
destroys the model's ability to express appropriate uncertainty, causing 
the UI $\rightarrow$ CI migration that degrades CVS even as accuracy 
stabilizes.

The optimal $\tau$ can be characterized in CVS terms: it should maximize 
CVS = CommitAcc $\times$ AppropUncert, not raw accuracy. A model with 
$\tau = 0.8$ that achieves 82\% accuracy but CVS = 0.52 is preferable 
to one with $\tau = 0.1$ achieving 83\% accuracy but CVS = 0.36.

This provides a principled answer to ``what temperature should Gumbel-Softmax use?'': the temperature that maximizes the Certainty-Validity Score, which balances reliable commitment against appropriate uncertainty. For our architecture and tasks, this empirically falls in the 0.7--0.9 range.

\paragraph{Architecture Specificity.} It is important to note that these dynamics---specifically the geometric resonance and refusal to commit to ambiguous signal---are intrinsic properties of the Eidos architecture's discrete commitment mechanism. They do not necessarily apply to standard continuous backpropagation models, which may exhibit different convergence behaviors (such as texture memorization) under similar conditions. The Certainty-Validity framework, however, remains applicable to any system capable of expressing uncertainty.

\section{Conclusion}

The 83\% accuracy ceiling observed across Fashion-MNIST, EMNIST, and IMDB 
reflects dataset ambiguity, not architectural limitation. Removing 
structurally ambiguous classes from Fashion-MNIST raises accuracy to 97\%. 
Filtering IMDB for strong sentiment reaches 87\%. EMNIST digits achieve 
99.59\%. In each case, filtering restores the Platonic Spike---the early 
positive generalization gap indicating structural discovery.

Approximately 17\% of samples in standard benchmarks are structurally 
ambiguous, requiring texture or context beyond pure topology. The 
architecture correctly refuses to memorize these samples, producing the 
83\% ceiling. This is not failure but principled behavior: commitment 
requires sufficient structural evidence.

The contribution is methodological as much as empirical. Standard metrics 
like accuracy, AUROC, precision, and recall are inadequate for evaluating 
discrete selection models because they treat all errors as equivalent 
failures. The Certainty-Validity framework we propose distinguishes 
confident errors (overconfidence, the true failure mode) from uncertain 
errors (appropriate abstention on ambiguous inputs). The derived metrics---Commitment 
Accuracy, Appropriate Uncertainty Rate, Coverage, and Certainty-Validity 
Score---provide a more complete characterization of model behavior than 
accuracy alone.

Ablation by ambiguity removal, combined with Platonic Spike detection and 
Certainty-Validity analysis, offers a methodology for distinguishing 
architectural capability from dataset ambiguity. Benchmark accuracy alone 
conflates these factors; the tools presented here separate them.

The 83\% ambiguity ceiling, far from representing failure, reveals the 
architecture working exactly as intended: learning structure where 
structure exists, committing confidently when evidence suffices, and 
withholding commitment where evidence is insufficient.

We conclude with a formal definition of \emph{Benign Overfitting} for discrete systems: it is the point where the Certainty-Validity Score (CVS) begins to decline, even if Test Accuracy remains stable or increases. This decline signals a pathological migration from Uncertain-Incorrect (valid epistemic doubt) to Confident-Incorrect (invalid hallucination). Therefore, the \textbf{Optimal Model State} is not simply the point of maximum accuracy, but the checkpoint where both Accuracy and CVS are maximized simultaneously. Trustworthiness, measured by CVS, is the true limit of model validity.

\section*{Supplementary Material}
This work is supported by open-source reproducibility artifacts, including training logs, model checkpoints, and experimental scripts \cite{anderson2026invariant}.

\bibliographystyle{plain}
\bibliography{references}

\end{document}